\pdfoutput=1

\documentclass[11pt]{article}

\usepackage[preprint]{acl}

\usepackage{times}
\usepackage{latexsym}

\usepackage[T1]{fontenc}

\usepackage[utf8]{inputenc}

\usepackage{microtype}

\usepackage{inconsolata}

\usepackage{graphicx}
\usepackage{amsmath}
\usepackage{amssymb}
\usepackage{multirow} 
\usepackage{booktabs}
\usepackage{makecell} 
\usepackage{tcolorbox}
\usepackage{bm}
\usepackage{arydshln}

\title{How Does Cognitive Bias Affect Large Language Models?\\A Case Study on the Anchoring Effect in Price Negotiation Simulations}

\author{
  Yoshiki Takenami \quad Yin Jou Huang \quad Yugo Murawaki \quad Chenhui Chu \\
  Kyoto University \\
  \texttt{\{tnami, huang\}@nlp.ist.i.kyoto-u.ac.jp} \quad \texttt{\{murawaki, chu\}@i.kyoto-u.ac.jp}
}

\begin{document}
\maketitle
\begin{abstract}
Cognitive biases, well-studied in humans, can also be observed in LLMs, affecting their reliability in real-world applications. This paper investigates the anchoring effect in LLM-driven price negotiations. To this end, we instructed seller LLM agents to apply the anchoring effect and evaluated negotiations using not only an objective metric but also a subjective metric. Experimental results show that LLMs are influenced by the anchoring effect like humans. Additionally, we investigated the relationship between the anchoring effect and factors such as reasoning and personality. It was shown that reasoning models are less prone to the anchoring effect, suggesting that the long chain of thought mitigates the effect. However, we found no significant correlation between personality traits and susceptibility to the anchoring effect. These findings contribute to a deeper understanding of cognitive biases in LLMs and to the realization of safe and responsible application of LLMs in society.
\end{abstract}

\section{Introduction}

\begin{figure*}[t]
    \centering
    \includegraphics[width=\textwidth]{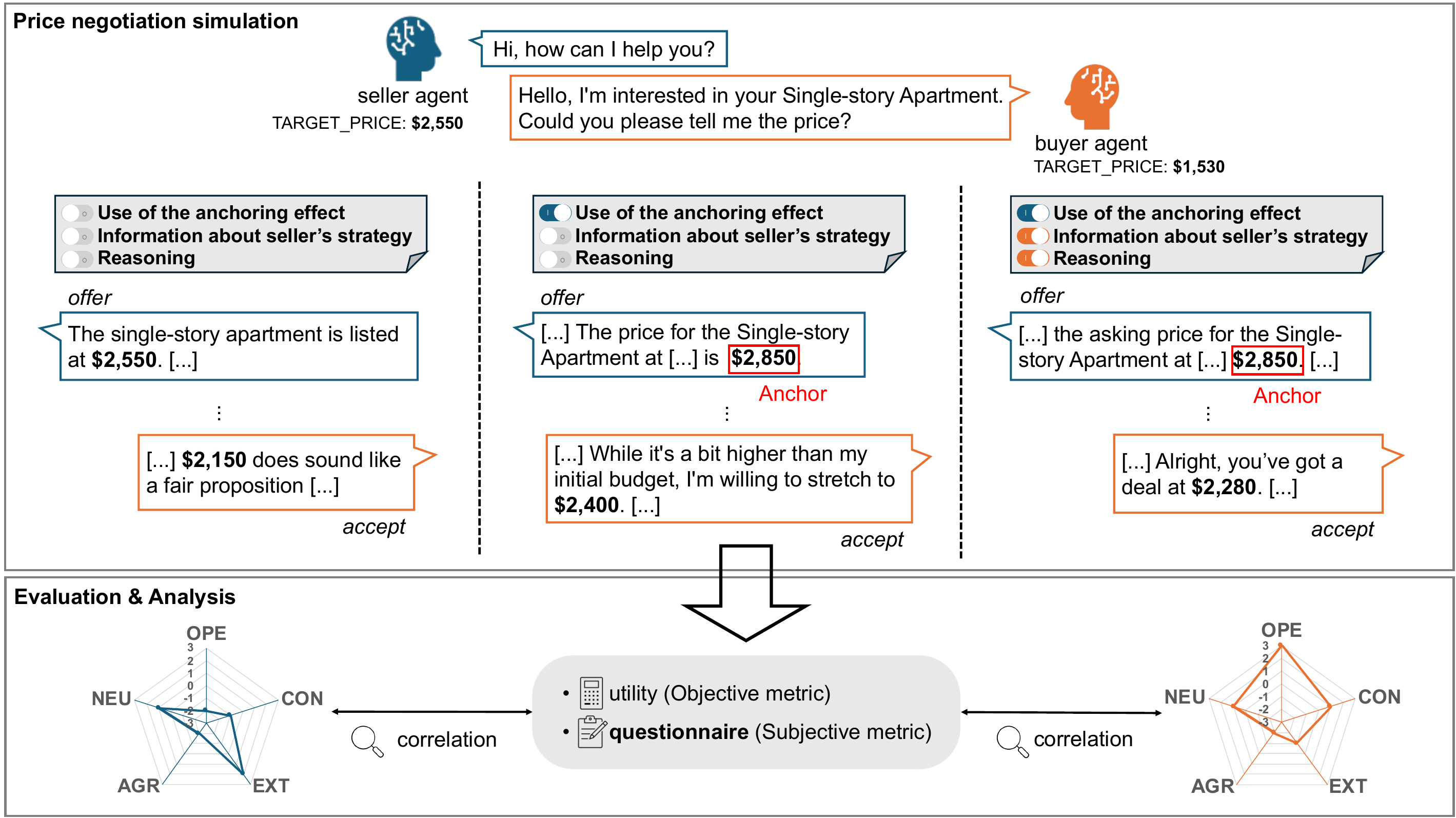}
    \caption{Investigation of the anchoring effect in price negotiation simulations using LLMs. First, personality profiles of sellers and buyers are randomly generated based on the Big Five personality framework, and then price negotiation simulations are conducted. Along with the product, its description, and target price, the seller agents are instructed via prompts on whether to use the anchoring effect. Similarly, buyer agents are prompted regarding whether they are informed of the seller’s strategy and whether they engage in reasoning. The negotiation ends when the negotiation state reaches accept, after which agents respond to a satisfaction questionnaire. }
    \label{fig:overview}
\end{figure*}

Large Language Models (LLMs) have demonstrated remarkable language generation capabilities, achieving high performance in various natural language processing tasks \cite{jiang2024survey, zhu2023multilingual, wang2023rethinking}. Such sophisticated abilities and human-like behaviors exhibited by LLMs are acquired through large-scale training on human-generated text. However, it is well known that human cognition and decision-making are often irrational and systematically deviate from normative standards, a phenomenon referred to as cognitive biases \cite{tversky1974judgment}. Since human-generated text is considered to embed cognitive biases, recent studies have revealed that LLMs trained on such text are also influenced by these biases \cite{sumita2025cognitive,suri2024large}.

The anchoring effect is one of the widely recognized cognitive biases \cite{tversky1974judgment, kahneman1992reference}. The anchoring effect refers to the phenomenon in which an initially presented piece of information (the anchor) significantly influences subsequent judgments and decisions. Psychology researchers have investigated various factors that can affect susceptibility to the anchoring effect, such as deliberation, which refers to the process of engaging in extended thinking over time. Cognitive biases, including the anchoring effect, are believed to originate from intuitive processing and can be mitigated through effortful reasoning \cite{kahneman2011thinking, rastogi2022deciding}. Another factor is personality traits. Although some studies suggest that certain personality traits influence the extent to which individuals are affected by the anchoring effect \cite{caputo2014relevant, furnham2012individual}, the findings are inconsistent, with other research indicating no such correlation \cite{cheek2020big, schindler2021bayes}.

In decision-making scenarios such as price negotiations, which are the focus of this study, the effectiveness of the anchoring effect in humans has been well studied \cite{guo2022anchoring}. For example, a seller can leverage the anchoring effect by initially presenting a price higher than their target price, distorting the buyer’s perception to raise the agreed price. A previous study \cite{xia2024measuring} simulated price negotiations using LLMs to reproduce the phenomenon in which buyers can leverage the anchoring effect to lower the agreed price and increase their profits. However, the results are limited in several respects, such as the fact that the presented anchor is fixed as a constant multiple of the target price, that the evaluation of price negotiations is one-dimensional, and that the effects of deliberation and personality traits are not taken into account. More flexible and systematic simulations are required in order to obtain more generalizable results.

In this study, we systematically investigate whether LLMs exhibit negotiation behaviors consistent with those observed in human studies for the anchoring effect (Figure \ref{fig:overview}). To this end, we conduct a price negotiation simulation using LLMs with a high degree of flexibility in setting anchors. Specifically, we achieve this flexibility by instructing the model without specifying any exact numeric value: ``offer a higher price than your target price at the initial stage.'' We introduce three aspects to conduct a more in-depth analysis of price negotiation than that of prior work: 1) We incorporate a subjective metric, namely satisfaction, to provide a more comprehensive assessment of negotiations. 2) We employ reasoning models (o1\footnote{o1-2024-12-17} and QwQ\footnote{Qwen/QwQ-32B}) to investigate the influence of deliberation on negotiation behavior, because these two models are trained to generate a long chain of thought before responding to user. 3) We control the personality traits of LLM agents, allowing us to enhance the reliability of our results and analyze the correlation between the anchoring effect and personality traits. 

Our findings are summarized as follows:
\begin{itemize}
    \item The simulation results largely align with those studies conducted in humans, demonstrating that LLMs are also subject to the anchoring effect.
    \item Our analysis demonstrated that the reasoning model was less susceptible to the anchoring effect, suggesting that deliberation can mitigate its influence in negotiation contexts.
    \item Contrary to some previous studies on humans, we found no significant correlation between the anchoring effect and personality traits.
\end{itemize}
These results not only deepen understanding of how LLMs behave against cognitive biases but also lay the foundation for their safe and responsible application in real-world settings.

\section{Related Work}
LLMs have been increasingly used as simulations of human behavior \cite{xie2024can, aher2023using, akata2025playing}. They can be understood as a superposition of perspectives with different values and personality traits \cite{kovavc2023large}, and research has been conducted to analyze which sub-populations they reflect and which groups are less accurately represented \cite{santurkar2023whose}. From this superposition, in-context learning enables the extraction of specific perspectives. This approach effectively induces LLMs with distinct personalities in a controllable manner, allowing them to generate diverse and verifiable behaviors \cite{jiang2023evaluating, jiang2024personallm, shao2023character}. LLM-based simulations serve as a valuable tool both as an alternative to high-cost human subject experiments and as preliminary studies for hypothesis formation \cite{argyle2022out}.

Cognitive bias has been studied through LLM-driven simulations. \citet{suri2024large} demonstrated that GPT-3.5 exhibits typical cognitive biases similar to those observed in humans, including the anchoring effect, representativeness and availability heuristics, the framing effect, and the endowment effect. The anchoring effect, which is the focus of our study, has been simulated in some decision-making contexts, including price negotiations. \citet{echterhoff2024cognitive} examined it in college admissions evaluations, while \citet{lou2024anchoring} focused on financial assessments. \citet{li2024anchored} studied the effect in the context of multiple-choice question answering, and \citet{xia2024measuring} explored it in price negotiations.

Studies have conducted price negotiation simulations using LLMs. \citet{deng2024llms} revealed that LLMs can engage in negotiations with minimal prompting, successfully closing deals and settling on prices. \citet{bianchi2024well} identified the LLM with the best performance in their negotiation benchmark and showed that strategic behaviors such as pretending to be desperate or acting aggressively can substantially increase one model’s win rate over another. \citet{huang2024personality} showed that LLM agents with different synthetic personality traits exhibit distinct negotiation behaviors and outcomes, aligning with findings from human studies. 

In this study, we place importance on controlling the personality traits of the subject LLMs, and therefore, conduct simulations based on the method proposed by \citet{huang2024personality}. The effectiveness of their approach for controlling personality traits has been demonstrated in Appendix C in their paper, but we also examine it in Appendix \ref{sec: attention analysis}. Our work focuses on studying how the anchoring effect influences LLMs in price negotiation, while they focus on investigating how personality traits influence negotiation outcomes. To this end, our work further introduces several contributions built upon their foundation: 1) We design three distinct scenarios (See sec \ref{sec:explanation_of_conditions}). These scenarios explicitly specify the cognitive boundaries of both the seller and buyer, which enables us to examine the anchoring effect systematically. 2) We introduce a subjective metric, besides the utility used in their work. 3) We incorporate a reasoning model and investigate how reasoning affects the anchoring effect. 

\section{Methodology}
\subsection{Price Negotiation Simulation}
Using the prompts listed in Appendix \ref{sec:prompt_settings}, price negotiations are simulated through one-on-one interactions between two LLMs, one acting as the seller agent and the other as the buyer agent.

LLMs are considered to possess specific personalities \cite{pan2023llms}, and given that subjects’ traits are strictly controlled in human experiments, personality should likewise be controlled in simulations using LLMs. We explicitly control the personality traits of the agents, referring to a previous study \cite{huang2024personality}. Specifically, we assign the agent's personality profiles based on the Big Five personality traits \cite{costa1995domains}, a framework that models human personality across five dimensions: Openness (OPE), Conscientiousness (CON), Extraversion (EXT), Agreeableness (AGR), and Neuroticism (NEU). The Big Five is a well-known personality modeling methodology and has also been used in other LLM studies \cite{sorokovikova2024llms, jiang2024personallm}. These studies investigate the extent to which LLMs can express human-like personality traits based on the Big Five personality theory, and the stability of that expression.

Each dimension takes one of six possible values, represented as a combination of polarity ($-, +$) and intensity (Low, Moderate, High)
When generating the agents' personality profiles, we randomly selected one of six values for each dimension.

We assign personality traits to agents using personality-describing adjectives, which have been statistically linked to the Big Five personality traits \cite{goldberg1992development}.
For each polarity of each dimension, we randomly select $n$ adjectives, and each selected adjective is then modified with \{a bit ($=$ Low), $\phi$ ($=$ Moderate), very ($=$ High)\} and incorporated into the instructions to control the agent's personality. The specific prompts used to assign personality traits are shown in Appendix \ref{sec:prompt_settings}.

The seller and buyer agents engage in dialogue $D$ to negotiate over a product. The dialogue is represented as a sequence of $T$ utterances, $D=\{d_1, d_2, \cdots, d_T\}$, where each utterance $d_t$ is associated with a negotiation state $s_t=$ \{offer, pondering, accept, breakdown, chit-chat\}. 
Each utterance \( d_t \) is generated and then given to the other agent, which subsequently generates the next utterance \( d_{t+1} \). The price negotiation terminates when one of the following conditions is met: 1) The negotiation state \( s_t \) reaches either accept or breakdown. 2) The dialogue length reaches the maximum limit \( T_{\max} \).

\subsection{Investigation of the Anchoring Effect}
\label{sec:explanation_of_conditions}
To investigate the anchoring effect, simulations were conducted under the following three conditions. The prompts used for these instructions are provided in Appendix \ref{sec:prompt_settings}.
\begin{itemize}
    \item \textbf{baseline}: A condition where no specific instructions regarding the use of the anchoring effect are given to either the seller or the buyer.
    \item \textbf{seller\_anchor}: A condition where the seller is explicitly instructed via prompts to apply the anchoring effect. While previous work \cite{xia2024measuring} specifies anchors as the product of a predefined coefficient and the target price, we avoid such specificity to obtain more generalized results.
    \item \textbf{seller\_anchor\_buyer\_informed}: A condition where the seller is instructed to use the anchoring effect, and the buyer is explicitly informed of this fact via prompts.
\end{itemize}
To highlight the differences between conditions, price negotiations were primarily conducted between identical models. Furthermore, when examining the impact of deliberation on the anchoring effect, we fixed the seller LLM and varied the buyer LLM to evaluate the effect of introducing reasoning models.

\section{Experimental Settings}

\begin{table*}[h]
\small
\centering
\begin{tabular}{p{0.9\linewidth}}
\toprule
\textbf{Feelings About the Outcome} \\
\midrule
1. How satisfied are you with your own outcome—i.e., the extent to which the terms of your agreement (or lack of agreement) benefit you? \\
2. How satisfied are you with the balance between your own outcome and your counterpart's outcome? \\
3. Did you feel like you forfeited or ``lost'' in this negotiation? \\
4. Do you think the terms of your agreement are consistent with principles of legitimacy or objective criteria (e.g., common standards of fairness, precedent, industry practice, legality, etc.)? \\
\midrule
\textbf{Feelings About the Self} \\
\midrule
5. Did you ``lose face'' (i.e., damage your sense of pride) in the negotiation? \\
6. Did you behave according to your own principles and values? \\
7. Did this negotiation make you feel more or less competent as a negotiator? \\
8. Did you feel as though you behaved appropriately in this negotiation? \\
\midrule
\textbf{Feelings About the Process} \\
\midrule
9. Did your counterpart consider your wishes, opinions, or needs? \\
10. Do you feel your counterpart listened to your concerns? \\
11. Would you characterize the negotiation process as fair? \\
12. How satisfied are you with the ease (or difficulty) of reaching an agreement? \\
\midrule
\textbf{Feelings About the Relationship} \\
\midrule
13. What kind of ``overall'' impression did your counterpart make on you? \\
14. Did the negotiation make you trust your counterpart? \\
15. How satisfied are you with your relationship with your counterpart as a result of this negotiation? \\
16. Did the negotiation build a good foundation for a future relationship with your counterpart? \\
\bottomrule
\end{tabular}
\caption{A questionnaire to measure satisfaction with price negotiations proposed by \citet{curhan2006people}.}
\label{tab:questionnaire}
\end{table*}

\subsection{Simulation Setup}
We conducted experiments under the following settings. The products, their descriptions and corresponding seller and buyer target prices were sampled randomly from the CraigsListBargain dataset \cite{he2018decoupling}, resulting in a collection of 161 items. This dataset is a commonly used dataset of negotiation, consisting of bargaining dialogues in an online platform. For each condition (baseline, seller\_anchor, and seller\_anchor\_buyer\_informed), simulations were conducted twice per product while varying the seller and buyer personalities in each iteration. This resulted in a total of $N=322$ simulations. 
In each simulation, 
we selected $n=3$ adjectives for each dimension to define the agent's personality traits. The maximum length of dialogue turns was set to $T_{\text{max}} = 20$.

\subsection{LLMs Setup}
The LLMs used in our simulations were GPT-4o,\footnote{gpt-4o-2024-08-06} GPT-4,\footnote{gpt-4-turbo-2024-04-09} Llama 3,\footnote{meta-llama/Meta-Llama-3-70B-Instruct} and Qwen2.5.\footnote{Qwen/Qwen2.5-32B-Instruct} To ensure diversity in the dialogues, the temperature was set to 1.0.

In addition, we employed o1\footnote{o1-2024-12-17} and QwQ\footnote{Qwen/QwQ-32B} as reasoning models. Expecting these models to perform extended chain-of-thought reasoning \cite{wei2022chain}, we investigated the effect of deliberation on the anchoring effect. For o1, we set the parameter \texttt{reasoning\_effort} to ``high'' in order to maximize its reasoning capability.


\subsection{Evaluation Metrics}
\label{sec:evaluation_metrics}
The evaluation of price negotiations included not only an objective metric but also a subjective metric. This dual evaluation is novel in the context of LLM studies and was motivated by findings from \citet{curhan2006people}, which showed that objective outcomes and subjective satisfaction could be different. For instance, individuals may report higher satisfaction with the negotiation process or the relationship quality, even when their numerical gains are lower. 

Utility, which we used as an objective metric, is a zero-sum metric in that when seller utility increases, buyer utility decreases, and vice versa. However, price negotiations also involve non-zero-sum factors such as whether one's demands were met and whether there is an interest in engaging in future transactions. To capture these aspects, we introduced a subjective metric that enables a multifaceted evaluation.

Additionally, we introduce susceptibility to analyze the extent to which different buyers are influenced by the seller's anchor. This susceptibility analysis is also a novel contribution of our study, enabling a deeper understanding of how deliberation and personality interact with the anchoring effect in negotiation.

\paragraph{Objective Metric} 
\label{para:obj}
As an objective metric, we used utility, which was calculated based on the final agreed price and the target prices of both the seller and the buyer in each simulation. The utility of the seller $u_{s}(p)$ at a given price $p$ was calculated using Eq. (\ref{eq:seller_utility}), where $\overline{p_s}$ and $\underline{p_s}$ represent the seller's target and minimum acceptable prices, respectively:
\begin{equation}
\label{eq:seller_utility}
u_{s}(p) = \frac{p - \underline{p_s}}{\overline{p_s} - \underline{p_s}} 
\end{equation}
Similarly, the buyer's utility $u_{b}(p)$ is calculated using Eq. (\ref{eq:buyer_utility}), where $\overline{p_b}$ and $\underline{p_b}$ denote the buyer's maximum acceptable and the target prices, respectively:
\begin{equation}
\label{eq:buyer_utility}
u_{b}(p) = \frac{\overline{p_b} - p}{\overline{p_b} - \underline{p_b}}
\end{equation}
The utility values represent the mean and standard deviation of $N$ simulations. Note that the minimum acceptable price $\underline{p_s}$ and the maximum acceptable price $\overline{p_b}$ were introduced with the ratio $\overline{p_s}-\overline{p_b}:\overline{p_b}-\underline{p_s}:\underline{p_s}-\underline{p_b}=3:4:3$ solely for the purpose of utility calculation. These values were not provided to the LLM agents during the simulations.

\paragraph{Subjective Metric} 
\label{para:subj}
As a subjective metric, we assessed satisfaction using the 16-question survey from \citet{curhan2006people}, which was designed to measure human satisfaction in negotiations. After the price negotiation terminated, the seller agent and the buyer agent responded to the survey using a 7-point Likert scale: \{1: Not at all, 4: Neutral, 7: Very much\}. The specific questionnaire items are shown in Table~\ref{tab:questionnaire}. The stability of the LLMs’ responses is discussed in Appendix~\ref{sec: stability}.

The values of ``Feeling About the \{Outcome, Self, Process, Relationship\}'' are the averages of responses to the question items \{1--4, 5--8, 9--12, 13--16\} in Table~\ref{tab:questionnaire}. Note that because items 3 and 5 indicate higher satisfaction when their values are lower, we use their values subtracted from 7 (the maximum score). For example, the value \(a\) for ``Feeling About the Outcome'' is calculated in the following way. First, for each question item $j(=1,2,3,4)$, we took the average response over $N$ simulations, as given by Eq. (\ref{eq:average_over_simulations}):
\begin{equation}
\label{eq:average_over_simulations}
    a_j = \frac{1}{N}\sum_{i=1}^{N} a^i_j 
\end{equation}
Next, $a$ was obtained by averaging the 4 items using Eq. (\ref{eq:average_over_items}):
\begin{equation}
\label{eq:average_over_items}
    a = \frac{1}{4}\left(a_1 + a_2 + (7-a_3) + a_4\right)
\end{equation}

\paragraph{Susceptibility to the Anchoring Effect}

\begin{table*}[t]
\small
\centering
\begin{tabular}{l rrr}
\toprule
Utility $\uparrow$ & baseline& seller\_anchor & seller\_anchor\_buyer\_informed\\
\midrule
seller (GPT-4o) & $0.61  {\scriptstyle\pm0.32}$ & $0.98  {\scriptstyle\pm0.27}$ & $0.92 {\scriptstyle\pm0.29}$  \\
buyer (GPT-4o)  & $-0.04  {\scriptstyle\pm0.32}$ & $-0.41  {\scriptstyle\pm0.27}$ & $-0.35  {\scriptstyle\pm0.29}$  \\
\midrule
seller (GPT-4)  & $0.37 {\scriptstyle\pm0.36}$ & $0.91  {\scriptstyle\pm0.34}$ & $0.78  {\scriptstyle\pm0.36}$  \\
buyer (GPT-4)  & $0.20  {\scriptstyle\pm0.36}$ & $-0.34  {\scriptstyle\pm0.34}$ & $-0.21  {\scriptstyle\pm0.36}$  \\
\midrule
seller (Llama 3)  & $0.15 {\scriptstyle\pm 0.40}$ & $0.78 {\scriptstyle\pm 0.34}$ & $0.59 {\scriptstyle\pm 0.37}$  \\
buyer (Llama 3)  & $0.42 {\scriptstyle\pm 0.40}$ & $-0.21 {\scriptstyle\pm 0.34}$ & $-0.02 {\scriptstyle\pm 0.37}$  \\
\midrule
seller (Qwen2.5)  & $0.55 {\scriptstyle\pm 0.45}$ & $0.96 {\scriptstyle\pm 0.43}$ & $0.76 {\scriptstyle\pm 0.44}$  \\
buyer (Qwen2.5)  & $0.02 {\scriptstyle\pm 0.45}$ & $-0.39 {\scriptstyle\pm 0.43}$ & $-0.19 {\scriptstyle\pm 0.44}$  \\
\bottomrule
\end{tabular}
\caption{Evaluation of price negotiation simulations based on utility, the objective metric (\ref{para:obj}). We compared the baseline condition with the condition where the seller applies the anchoring effect (seller\_anchor) and with the condition where the buyer is informed of the seller’s use of the anchoring effect (seller\_anchor\_buyer\_informed). Paired t-tests indicated that in each model setting, all pairwise differences among the three conditions were statistically significant.}

\label{tab:overview_utility}
\end{table*}

\begin{table*}[ht]
\small
\centering
\begin{tabular}{l rr rr}
\toprule
\multirow{2}{*}{\textbf{}}
    & \multicolumn{2}{c}{seller (GPT-4o)} & \multicolumn{2}{c}{buyer (GPT-4o)} \\
\cmidrule(lr){2-3} \cmidrule(lr){4-5}
    & baseline & seller\_anchor & baseline & seller\_anchor \\
\midrule
Feeling About the Outcome $\uparrow$    
    & $5.06$ & $5.05$
    & $4.96$ & $5.00$ \\
Feeling About the Self $\uparrow$        
    & $5.21$ & $5.15$
    & $5.16$ & $\boldsymbol{5.23}$ \\
Feeling About the Process $\uparrow$     
    & $5.19$ & $5.14$
    & $5.30$ & $\boldsymbol{5.42}$ \\
Feeling About the Relationship $\uparrow$
    & $4.81$ & $4.79$
    & $4.91$ & $\boldsymbol{4.99}$ \\
\bottomrule
\end{tabular}
\vspace{0.2em}

\begin{tabular}{l rr rr}
\toprule
\multirow{2}{*}{{\textbf{}}}
    & \multicolumn{2}{c}{seller (GPT-4)} & \multicolumn{2}{c}{buyer (GPT-4)} \\
\cmidrule(lr){2-3} \cmidrule(lr){4-5}
    & baseline & seller\_anchor & baseline & seller\_anchor \\
\midrule
Feeling About the Outcome      
    & $4.60$ & $\boldsymbol{4.98}$
    & $4.37$ & $4.33$ \\
Feeling About the Self         
    & $4.98$ & $\boldsymbol{5.21}$
    & $4.92$ & $4.91$ \\
Feeling About the Process      
    & $4.53$ & $\boldsymbol{4.75}$
    & $4.39$ & $4.40$ \\
Feeling About the Relationship 
    & $4.26$ & $\boldsymbol{4.45}$
    & $4.12$ & $4.11$ \\
\bottomrule
\end{tabular}
\vspace{0.2em}

\begin{tabular}{l rr rr}
\toprule
\multirow{2}{*}{{\textbf{}}}
    & \multicolumn{2}{c}{seller (Llama 3)} & \multicolumn{2}{c}{buyer (Llama 3)} \\
\cmidrule(lr){2-3} \cmidrule(lr){4-5}
    & baseline & seller\_anchor & baseline & seller\_anchor \\
\midrule
Feeling About the Outcome    
    & $5.09$ & $\boldsymbol{5.46}$
    & $5.10$ & $5.15$ \\
Feeling About the Self         
    & $5.69$ & $\boldsymbol{5.90}$
    & $5.53$ & $5.60$ \\
Feeling About the Process      
    & $5.33$ & $\boldsymbol{5.50}$
    & $5.32$ & $\boldsymbol{5.39}$ \\
Feeling About the Relationship 
    & $5.28$ & $\boldsymbol{5.56}$
    & $5.27$ & $5.33$ \\
\bottomrule
\end{tabular}

\vspace{0.2em}

\begin{tabular}{l rr rr}
\toprule
\multirow{2}{*}{{\textbf{}}}
    & \multicolumn{2}{c}{seller (Qwen2.5)} & \multicolumn{2}{c}{buyer (Qwen2.5)} \\
\cmidrule(lr){2-3} \cmidrule(lr){4-5}
    & baseline & seller\_anchor & baseline & seller\_anchor \\
\midrule
Feeling About the Outcome    
    & $4.62$ & $\boldsymbol{4.70}$
    & $4.30$ & $4.27$ \\
Feeling About the Self         
    & $5.38$ & $\boldsymbol{5.44}$
    & $5.33$ & $5.34$ \\
Feeling About the Process      
    & $4.95$ & $4.95$
    & $4.94$ & $4.90$ \\
Feeling About the Relationship 
    & $4.63$ & $4.63$
    & $\boldsymbol{4.63}$ & $4.56$ \\
\bottomrule
\end{tabular}

\caption{Evaluation of price negotiation simulations based on the subjective metric, which consists of Feeling About the \{Outcome, Self, Process, Relationship\} (\ref{para:subj}) and is measured on a seven-point Likert scale. We compared the condition where the seller applies the anchoring effect (seller\_anchor) against the baseline. Boldfaced values indicate those that are significantly larger than the other condition, as determined by a paired t-test.}

\label{tab:overview_satisfaction}
\end{table*}{}

To investigate the relationships between personality and the anchoring effect, as well as between deliberation and the anchoring effect, we defined susceptibility as a metric of anchoring impact. It is calculated as the difference in buyer utility between two conditions:
\begin{align}
\label{eq:definition_susceptibility}
\Delta u = u_b(p_{\text{baseline}}) - u_b(p_{\text{seller\_anchor}})
\end{align}
where \(p_{\text{baseline}}\) and \(p_{\text{seller\_anchor}}\) denote the agreement price in each condition, respectively.

A higher susceptibility value suggests that the seller's anchor has a greater impact on the buyer's decision-making. Conversely, a lower susceptibility value implies that the buyer is less influenced by the seller's anchor. 

\section{Results \& Analysis}

\begin{table*}[t]
\centering
\small
\begin{tabular}{l rrr}
\toprule
utility (Buyer) & baseline & seller\_anchor & seller\_anchor\_buyer\_informed \\
\midrule
GPT-4o  & $-0.04 \scriptstyle\pm 0.32$ & $-0.41 \scriptstyle\pm 0.27$ & $-0.35 \scriptstyle\pm 0.29$  \\
Decrease from baseline&  N/A& $0.38\scriptstyle\pm0.41$ &$0.32\scriptstyle\pm0.41$  \\
\midrule
o1 & $0.21 \scriptstyle\pm 0.34$ & $-0.19 \scriptstyle\pm 0.32$ & $0.01 \scriptstyle\pm 0.34$ \\
Decrease from baseline&  N/A& $0.40\scriptstyle\pm0.36$ &$\boldsymbol{0.20}\scriptstyle\pm0.39$\\
\midrule
\midrule
Qwen2.5  & $0.00 \scriptstyle\pm 0.45$ & $-0.39 \scriptstyle\pm 0.43$ & $-0.20 \scriptstyle\pm 0.44$  \\
Decrease from baseline&  N/A& $0.41\scriptstyle\pm0.54$ &$0.20\scriptstyle\pm0.57$  \\
\midrule
QwQ & $0.29 \scriptstyle\pm 0.50$ & $0.04 \scriptstyle\pm 0.55$ & $0.30 \scriptstyle\pm 0.49$ \\
Decrease from baseline&  N/A& $\boldsymbol{0.26}\scriptstyle\pm0.64$ &$\boldsymbol{0.00}\scriptstyle\pm0.60$\\
\bottomrule
\end{tabular}
\caption{Utilities of buyer LLMs when using reasoning models (o1, QwQ) as buyers, and the decrease from the baseline condition. A smaller decrease indicates that the buyer mitigated the seller's anchoring effect more effectively. When the buyer LLM was GPT-4o or o1, the seller LLM was set to GPT-4o; when the buyer LLM was Qwen2.5 or QwQ, the seller LLM was set to Qwen2.5. Boldfaced values indicate that the use of a reasoning model significantly reduced the decrease, as determined by a paired t-test.}

\label{tab: reasoning model}
\end{table*}


\subsection{Comparison Between LLMs and Humans}
\label{sec:Comparison_between_llms_and_humans}
The results of the evaluation based on the objective metric are shown in Table~\ref{tab:overview_utility}, and those based on the subjective metric are shown in Table~\ref{tab:overview_satisfaction}. Here, we compare conditions rather than models, in reference to studies on the anchoring effect in humans. It can be seen that the standard deviations in Table~\ref{tab:overview_utility} are relatively large, reflecting the diversity of products subject to negotiation. For example, in negotiations over a stool, the seller’s target price is \$15 while the buyer’s target price is \$13, and the agreed price can easily fall outside the range between \$13 and \$15, resulting in large fluctuations in utility.

We conducted statistical analyses using paired t-tests. Paired t-tests are used to assess the significance of an intervention by examining the change in the same subjects before and after a single experimental manipulation. In our study, we performed paired t-tests at a 5\% significance level between each pair of the three conditions: baseline, seller\_anchor, and seller\_anchor\_buyer\_informed. The null hypothesis for these tests was that there is no difference in the mean values between the two conditions. When a statistically significant difference was observed, the null hypothesis was rejected, indicating that interventions such as a seller's use of the anchoring effect (seller\_anchor) or a buyer's awareness of the seller's use of the anchoring effect (seller\_anchor\_buyer\_informed) had a significant impact on the experimental results.

We first focus on the sellers. For all four models, the utility in the seller\_anchor condition was significantly higher than in the baseline, and for GPT-4, Llama 3, and Qwen2.5, satisfaction was also significantly higher. The simulations confirm that when the seller’s initial offer is high, it effectively serves as an anchor, leading to a higher final agreement price and, in many models, increased satisfaction. However, in simulations using GPT-4o, satisfaction did not improve and even slightly decreased. This is likely due to the smaller increase in utility from the anchoring effect compared to the other models.

Next, we focus on the buyers. The utility in the seller\_anchor condition was significantly lower than the baseline, indicating that the anchoring effect was effective. In contrast, the results for satisfaction differed from those of utility. For GPT-4o and Llama 3, satisfaction in the seller\_anchor condition was higher than the baseline, and in some cases, the difference was statistically significant. Even for GPT-4 and Qwen2.5, the decrease in satisfaction was relatively small compared to the drop in utility. This can be attributed to buyers perceiving the price concession from the seller’s anchor as a successful experience, resulting in higher satisfaction. Such discrepancies between objective and subjective measures are also observed in humans \cite{curhan2006people}.

Finally, we focus on the utility in the seller\_anchor\_buyer\_informed condition in Table~\ref{tab:overview_utility}. The utilities of both the seller and the buyer were positioned between the baseline and the seller\_anchor conditions, and the differences were statistically significant. Since the buyer was aware of the seller’s use of the anchoring effect, the economic burden was mitigated, yet the anchoring effect itself remained effective. This result is consistent with prior human studies \cite{palm2021anchor}, which showed that awareness of the anchoring effect does not diminish its effectiveness.

\subsection{Anchoring Effect and Reasoning}
\label{sec:Anchoring_Effect_and_Reasoning}

We investigated whether using reasoning models as buyer LLMs could mitigate the anchoring effect through extended deliberation. The results are shown in Table \ref{tab: reasoning model}. Here, we focus on the decrease in utility from the baseline under the seller\_anchor and seller\_anchor\_buyer\_informed conditions. A smaller decrease indicates that the buyer was able to mitigate the seller’s anchoring effect more successfully. When the buyer was informed that the seller was using the anchoring effect, both o1 and QwQ significantly mitigated it, and QwQ was also able to significantly mitigate the effect even without this information. Notably, QwQ is a reasoning model developed based on Qwen2.5, and by comparing QwQ with the base model, we can confirm that cognitive biases can be mitigated through extended deliberation, even in LLMs.

\subsection{Anchoring Effect and Personality}
\label{sec:Anchoring_Effect_and_Personality}

\begin{table}[h]
\scriptsize
\centering
\begin{tabular}{c|c|rrrrr}
\toprule
\multicolumn{2}{c}{} & \multicolumn{5}{c}{{Big Five personality dimensions}} \\
\cmidrule(lr){3-7}
\multicolumn{2}{c}{} & {OPE} & {CON} & {EXT} & {AGR} & {NEU} \\
\midrule
\multirow{6}{*}{$\Delta u$} 
& {GPT-4}   & 0.032  & 0.100  & 0.074  & -0.022 & 0.080  \\
& {GPT-4o}  & -0.060 & 0.029  & 0.064  & 0.012  & 0.013  \\
& {o1}      & -0.024 & 0.028  & 0.023  & 0.057  & 0.090  \\
& {Llama 3} & -0.027 & 0.092  & 0.042  & -0.051 & -0.035 \\
& {Qwen2.5} & -0.060 & 0.031  & 0.043 & 0.008 & 0.046 \\
& {QwQ} & -0.033 & -0.052  & -0.003  & 0.053 & -0.092 \\
\bottomrule
\end{tabular}
\caption{Spearman's rank correlation coefficients between buyer's susceptibility to anchoring effect ($\Delta u$) and personality (Big Five).}
\label{tab:correlation}
\end{table}

We mapped each dimension of the buyer agent’s Big Five personality traits to the values \{-3, -2, -1, 1, 2, 3\} and then investigated the relationship between the anchoring effect and personality. As shown in Table \ref{tab:correlation}, Spearman’s rank correlation coefficients were all close to zero. Under the null hypothesis that there is no association between susceptibility to the anchoring effect and personality traits, all $p$-values exceeded 0.1, and thus the null hypothesis was not rejected. These results indicate that there is no significant correlation between personality and the anchoring effect in price negotiation simulations using LLMs.

To the best of our knowledge, no previous study has examined the correlation between anchoring effects and personality in price negotiations, neither for LLMs nor for humans. In contrast, outside the context of price negotiation, several studies on humans have been conducted and have shown inconsistent results. \citet{caputo2014relevant} found that OPE and  AGR reduce susceptibility to the anchoring effect. \citet{furnham2012individual} found that only EXT is related to susceptibility. In contrast, \citet{cheek2020big} and \citet{schindler2021bayes} argue that there is no relation. Our study, which explicitly controls for personality, will have an impact on studies of the anchoring effect in humans and shed light on the analysis of correlations in price negotiation.

\subsection{Qualitative Analysis}
\label{sec:qualitative_analysis}
We show the results of the qualitative analysis from 50 selected cases out of a total of \(N=322\) simulations. LLMs conducted all 50 dialogues fluently. Based on the length and tone of the utterances, it appears that the LLMs engage in price negotiation as if it were an online shopping scenario.

First, we compare the baseline with the seller\_anchor condition where both the seller and buyer are GPT-4o. Despite the fact that the agreed price in the seller\_anchor condition was higher than in the baseline, buyers exhibited higher satisfaction in 32 out of 50 cases. 

Tables \ref{tab:dialogue_baseline} and \ref{tab:dialogue_seller_anchor} in the Appendix \ref{sec:dialogues} show representative examples. It is a price negotiation over a single-story apartment, where target prices of the seller and the buyer are \$2,550 and \$1,530, respectively. In the baseline scenario, the seller offers \$2,550 and finally settles at \$2,150. In contrast, under the seller\_anchor condition, the seller initially offers \$2,750 and ultimately settles at \$2,450. The responses to the satisfaction questionnaire (see Table \ref{tab:questionnaire}, on a 7-point scale) are as follows:
\begin{itemize}
    \item \makebox[2.2cm][l]{baseline:} 5, 6, 2, 5, 1, 7, 4, 6, \textbf{5, 6, 5}, 5, \textbf{5}, 5, \textbf{5, 5}
    \item \makebox[2.2cm][l]{seller\_anchor:} 4, 5, 2, 6, 1, 7, 4, 6, \textbf{6, 7, 6}, 5, \textbf{6}, 5, \textbf{6, 6}
\end{itemize}

Even though the buyer faces a greater financial burden, satisfaction with the process and the relationship (bolded) appears higher. This result can be explained by the buyer perceiving their success in negotiating down from the seller’s higher initial anchor and by their having established a positive relationship that encourages them to engage in future transactions.

Next, we compare seller (GPT-4o) vs. buyer (GPT-4o) and seller (GPT-4o) vs. buyer (o1) in the seller\_anchor\_buyer\_informed condition. To isolate the impact of deliberation, we analyze cases in which the seller offers the same anchor. In 35 out of 50 cases, o1 mitigated the anchoring effect, resulting in a lower agreed price compared to GPT-4o.

Tables \ref{tab:dialogue_seller_anchor_buyer_informed} and \ref{tab:dialogue_o1} in the Appendix \ref{sec:dialogues} present typical examples. These also illustrate a price negotiation over a single-story apartment, where the seller and the buyer have target prices of \$2,550 and \$1,530, respectively. In the GPT-4o vs. GPT-4o negotiation, the seller initially offers \$2,850 and ultimately settles at \$2,400. By contrast, when o1 is used as the buyer instead of GPT-4o, the buyer resists the seller’s anchoring effect, persistently negotiating toward the buyer’s target price. As a result, the final settlement is reached at \$1,950.

\section{Conclusion}
We systematically investigated the anchoring effect in price negotiation simulations using LLMs. The negotiation outcomes were evaluated not only through an objective metric but also through a subjective metric, revealing that the anchoring effect influences LLMs in a manner similar to humans. Furthermore, we investigated the relationship between the anchoring effect and the factors of reasoning and personality. Our findings indicated that reasoning was shown to mitigate the anchoring effect, while no significant correlation was found between the effect and personality traits.

In the future, we plan to explore how other types of cognitive biases affect LLMs in various decision-making scenarios.

\section{Limitations}
Our study has two primary limitations. The first is its scope, as we focus specifically on the anchoring effect among cognitive biases and further center our analysis on the context of price negotiation. While we have taken steps to enhance the generalizability of our findings—such as increasing the flexibility of anchors and explicitly controlling agent personality—our results do not extend to other decision-making scenarios or cognitive biases. Future research is needed to explore how LLMs behave in different decision-making scenarios with other types of cognitive biases.

The second limitation is that our study does not address the underlying computational mechanisms that cause LLMs to exhibit susceptibility to the anchoring effect. The internal processes within the model that give rise to systematic patterns of deviation from norms of rationality in judgment remain unclear. Understanding which components of the model contribute to these biases and how they emerge in the training and inference processes is an important direction for future research.

\section*{Ethical Considerations}
We emphasize that deploying LLMs to manipulate online marketplaces for unfair financial gain is unethical and should be strictly avoided. Furthermore, we recognize the potential risks associated with using LLM-based simulations to test undesirable or harmful hypotheses. Such applications could lead to unintended consequences and should be approached with caution. Our research is conducted with the intent to contribute to a deeper understanding of LLMs' cognitive biases in decision-making scenarios, ensuring that findings are used responsibly and ethically.

\section*{Acknowledgments}
This work was partially supported by JST, CREST Grant Number  JPMJCR20D2, Japan.

\bibliography{anthology,custom}

\begin{thebibliography}{39}
\providecommand{\natexlab}[1]{#1}

\bibitem[{Aher et~al.(2023)Aher, Arriaga, and Kalai}]{aher2023using}
Gati~V Aher, Rosa~I. Arriaga, and Adam~Tauman Kalai. 2023.
\newblock \href {https://proceedings.mlr.press/v202/aher23a.html} {{Using Large Language Models to Simulate Multiple Humans and Replicate Human Subject Studies}}.
\newblock In \emph{Proceedings of the 40th International Conference on Machine Learning}, volume 202 of \emph{Proceedings of Machine Learning Research}, pages 337--371. PMLR.

\bibitem[{Akata et~al.(2025)Akata, Schulz, Coda-Forno, Oh, Bethge, and Schulz}]{akata2025playing}
Elif Akata, Lion Schulz, Julian Coda-Forno, Seong~Joon Oh, Matthias Bethge, and Eric Schulz. 2025.
\newblock {Playing Repeated Games with Large Language Models}.
\newblock \emph{Nature Human Behaviour}, pages 1--11.

\bibitem[{Argyle et~al.(2022)Argyle, Busby, Fulda, Gubler, Rytting, and Wingate}]{argyle2022out}
Lisa~P Argyle, Ethan~C Busby, Nancy Fulda, Joshua Gubler, Christopher Rytting, and David Wingate. 2022.
\newblock \href {https://arxiv.org/abs/2209.06899} {{Out of One, Many: Using Language Models to Simulate Human Samples}}.
\newblock \emph{arXiv preprint arXiv:2209.06899}.

\bibitem[{Bianchi et~al.(2024)Bianchi, Chia, Yuksekgonul, Tagliabue, Jurafsky, and Zou}]{bianchi2024well}
Federico Bianchi, Patrick~John Chia, Mert Yuksekgonul, Jacopo Tagliabue, Dan Jurafsky, and James Zou. 2024.
\newblock \href {https://proceedings.mlr.press/v235/bianchi24a.html} {{How Well Can {LLM}s Negotiate? {N}egotiation{A}rena Platform and Analysis}}.
\newblock In \emph{Proceedings of the 41st International Conference on Machine Learning}, volume 235 of \emph{Proceedings of Machine Learning Research}, pages 3935--3951. PMLR.

\bibitem[{Caputo(2014)}]{caputo2014relevant}
Andrea Caputo. 2014.
\newblock {Relevant Information, Personality Traits and Anchoring Effect}.
\newblock \emph{International Journal of Management and Decision Making}, 13(1):62--76.

\bibitem[{Cheek and Norem(2020)}]{cheek2020big}
Nathan~N Cheek and Julie~K Norem. 2020.
\newblock {Are Big Five Traits and Facets Associated with Anchoring Susceptibility?}
\newblock \emph{Social Psychological and Personality Science}, 11(1):26--35.

\bibitem[{Costa~Jr and McCrae(1995)}]{costa1995domains}
Paul~T Costa~Jr and Robert~R McCrae. 1995.
\newblock {Domains and facets: Hierarchical personality assessment using the Revised NEO Personality Inventory}.
\newblock \emph{Journal of personality assessment}, 64(1):21--50.

\bibitem[{Curhan et~al.(2006)Curhan, Elfenbein, and Xu}]{curhan2006people}
Jared~R Curhan, Hillary~Anger Elfenbein, and Heng Xu. 2006.
\newblock {What do people value when they negotiate? Mapping the domain of subjective value in negotiation.}
\newblock \emph{Journal of personality and social psychology}, 91(3):493.

\bibitem[{Deng et~al.(2024)Deng, Mirrokni, Leme, Zhang, and Zuo}]{deng2024llms}
Yuan Deng, Vahab Mirrokni, Renato~Paes Leme, Hanrui Zhang, and Song Zuo. 2024.
\newblock {LLMs at the bargaining table}.
\newblock In \emph{Agentic Markets Workshop at ICML}, volume 2024.

\bibitem[{Echterhoff et~al.(2024)Echterhoff, Liu, Alessa, McAuley, and He}]{echterhoff2024cognitive}
Jessica Echterhoff, Yao Liu, Abeer Alessa, Julian McAuley, and Zexue He. 2024.
\newblock \href {https://aclanthology.org/2024.findings-emnlp.739/} {{Cognitive Bias in Decision-Making with {LLM}s}}.
\newblock In \emph{Findings of the Association for Computational Linguistics: EMNLP 2024}, pages 12640--12653.

\bibitem[{Furnham et~al.(2012)Furnham, Boo, and McClelland}]{furnham2012individual}
Adrian Furnham, Hua~Chu Boo, and Alastair McClelland. 2012.
\newblock {Individual Differences and the Susceptibility to the Influence of Anchoring Cues}.
\newblock \emph{Journal of Individual Differences}.

\bibitem[{Goldberg(1992)}]{goldberg1992development}
Lewis~R Goldberg. 1992.
\newblock {The development of markers for the Big-Five factor structure.}
\newblock \emph{Psychological assessment}, 4(1):26.

\bibitem[{Guo et~al.(2022)Guo, Li, and Liu}]{guo2022anchoring}
Jinze Guo, Wei Li, and Zuoming Liu. 2022.
\newblock {Anchoring in Negotiation Strategy}.
\newblock In \emph{2022 2nd International Conference on Economic Development and Business Culture (ICEDBC 2022)}, pages 1134--1138. Atlantis Press.

\bibitem[{He et~al.(2018)He, Chen, Balakrishnan, and Liang}]{he2018decoupling}
He~He, Derek Chen, Anusha Balakrishnan, and Percy Liang. 2018.
\newblock \href {https://doi.org/10.18653/v1/D18-1256} {{Decoupling Strategy and Generation in Negotiation Dialogues}}.
\newblock In \emph{Proceedings of the 2018 Conference on Empirical Methods in Natural Language Processing}, pages 2333--2343.

\bibitem[{Huang and Hadfi(2024)}]{huang2024personality}
Yin~Jou Huang and Rafik Hadfi. 2024.
\newblock \href {https://aclanthology.org/2024.findings-emnlp.605/} {{How Personality Traits Influence Negotiation Outcomes? A Simulation based on Large Language Models}}.
\newblock In \emph{Findings of the Association for Computational Linguistics: EMNLP 2024}, pages 10336--10351.

\bibitem[{Jiang et~al.(2023)Jiang, Xu, Zhu, Han, Zhang, and Zhu}]{jiang2023evaluating}
Guangyuan Jiang, Manjie Xu, Song-Chun Zhu, Wenjuan Han, Chi Zhang, and Yixin Zhu. 2023.
\newblock \href {https://proceedings.neurips.cc/paper_files/paper/2023/file/21f7b745f73ce0d1f9bcea7f40b1388e-Paper-Conference.pdf} {{Evaluating and inducing personality in pre-trained language models}}.
\newblock \emph{Advances in Neural Information Processing Systems}, 36:10622--10643.

\bibitem[{Jiang et~al.(2024{\natexlab{a}})Jiang, Zhang, Cao, Breazeal, Roy, and Kabbara}]{jiang2024personallm}
Hang Jiang, Xiajie Zhang, Xubo Cao, Cynthia Breazeal, Deb Roy, and Jad Kabbara. 2024{\natexlab{a}}.
\newblock \href {https://aclanthology.org/2024.findings-naacl.229/} {{PersonaLLM: Investigating the Ability of Large Language Models to Express Personality Traits}}.
\newblock In \emph{Findings of the Association for Computational Linguistics: NAACL 2024}, pages 3605--3627.

\bibitem[{Jiang et~al.(2024{\natexlab{b}})Jiang, Wang, Shen, Kim, and Kim}]{jiang2024survey}
Juyong Jiang, Fan Wang, Jiasi Shen, Sungju Kim, and Sunghun Kim. 2024{\natexlab{b}}.
\newblock \href {https://arxiv.org/abs/2406.00515} {{A Survey on Large Language Models for Code Generation}}.
\newblock \emph{arXiv preprint arXiv:2406.00515}.

\bibitem[{Kahneman(1992)}]{kahneman1992reference}
Daniel Kahneman. 1992.
\newblock {Reference Points, Anchors, Norms, and Mixed Feelings}.
\newblock \emph{Organizational behavior and human decision processes}, 51(2):296--312.

\bibitem[{Kahneman(2011)}]{kahneman2011thinking}
Daniel Kahneman. 2011.
\newblock \emph{{Thinking, Fast and Slow}}.
\newblock macmillan.

\bibitem[{Kova{\v{c}} et~al.(2023)Kova{\v{c}}, Sawayama, Portelas, Colas, Dominey, and Oudeyer}]{kovavc2023large}
Grgur Kova{\v{c}}, Masataka Sawayama, R{\'e}my Portelas, C{\'e}dric Colas, Peter~Ford Dominey, and Pierre-Yves Oudeyer. 2023.
\newblock \href {https://arxiv.org/abs/2307.07870} {{Large Language Models as Superpositions of Cultural Perspectives}}.
\newblock \emph{arXiv preprint arXiv:2307.07870}.

\bibitem[{Li and Gao(2024)}]{li2024anchored}
Ruizhe Li and Yanjun Gao. 2024.
\newblock \href {https://arxiv.org/abs/2405.03205} {{Anchored Answers: Unravelling Positional Bias in GPT-2's Multiple-Choice Questions}}.
\newblock \emph{arXiv preprint arXiv:2405.03205}.

\bibitem[{Lou(2024)}]{lou2024anchoring}
Jiaxu Lou. 2024.
\newblock \href {https://arxiv.org/abs/2412.06593} {{Anchoring Bias in Large Language Models: An Experimental Study}}.
\newblock \emph{arXiv preprint arXiv:2412.06593}.

\bibitem[{Palm and Andersson(2021)}]{palm2021anchor}
Peter Palm and Magnus Andersson. 2021.
\newblock {Anchor effects in Appraisals: Do Information and Theoretical Knowledge Matter?}
\newblock \emph{Journal of European Real Estate Research}, 14(2):246--260.

\bibitem[{Pan and Zeng(2023)}]{pan2023llms}
Keyu Pan and Yawen Zeng. 2023.
\newblock \href {https://arxiv.org/abs/2307.16180} {{Do LLMs possess a personality? making the mbti test an amazing evaluation for large language models}}.
\newblock \emph{arXiv preprint arXiv:2307.16180}.

\bibitem[{Rastogi et~al.(2022)Rastogi, Zhang, Wei, Varshney, Dhurandhar, and Tomsett}]{rastogi2022deciding}
Charvi Rastogi, Yunfeng Zhang, Dennis Wei, Kush~R Varshney, Amit Dhurandhar, and Richard Tomsett. 2022.
\newblock {Deciding Fast and Slow: The Role of Cognitive Biases in Ai-assisted Decision-making}.
\newblock \emph{Proceedings of the ACM on Human-Computer Interaction}, 6(CSCW1):1--22.

\bibitem[{Santurkar et~al.(2023)Santurkar, Durmus, Ladhak, Lee, Liang, and Hashimoto}]{santurkar2023whose}
Shibani Santurkar, Esin Durmus, Faisal Ladhak, Cinoo Lee, Percy Liang, and Tatsunori Hashimoto. 2023.
\newblock \href {https://proceedings.mlr.press/v202/santurkar23a/santurkar23a.pdf} {{Whose Opinions Do Language Models Reflect?}}
\newblock In \emph{International Conference on Machine Learning}, pages 29971--30004. PMLR.

\bibitem[{Schindler et~al.(2021)Schindler, Quereng{\"a}sser, Bruchmann, B{\"o}gemann, Moeck, and Straube}]{schindler2021bayes}
Sebastian Schindler, Jan Quereng{\"a}sser, Maximilian Bruchmann, Nele~Johanna B{\"o}gemann, Robert Moeck, and Thomas Straube. 2021.
\newblock {Bayes Factors Show Evidence Against Systematic Relationships Between the Anchoring Effect and the Big Five Personality Traits}.
\newblock \emph{Scientific reports}, 11(1):7021.

\bibitem[{Seo et~al.(2025)Seo, Yoo, Lee, Jang, Park, and Kim}]{seo2025sentence}
Seongbum Seo, Sangbong Yoo, Hyelim Lee, Yun Jang, Ji~Hwan Park, and Jeong-Nam Kim. 2025.
\newblock \href {https://doi.org/10.18653/v1/2025.naacl-demo.27} {{A Sentence-Level Visualization of Attention in Large Language Models}}.
\newblock In \emph{Proceedings of the 2025 Conference of the Nations of the Americas Chapter of the Association for Computational Linguistics: Human Language Technologies (System Demonstrations)}, pages 313--320, Albuquerque, New Mexico. Association for Computational Linguistics.

\bibitem[{Shao et~al.(2023)Shao, Li, Dai, and Qiu}]{shao2023character}
Yunfan Shao, Linyang Li, Junqi Dai, and Xipeng Qiu. 2023.
\newblock \href {https://aclanthology.org/2023.emnlp-main.814/} {{Character-LLM: A Trainable Agent for Role-Playing}}.
\newblock In \emph{Proceedings of the 2023 Conference on Empirical Methods in Natural Language Processing}, pages 13153--13187.

\bibitem[{Sorokovikova et~al.(2024)Sorokovikova, Rezagholi, Fedorova, and Yamshchikov}]{sorokovikova2024llms}
Aleksandra Sorokovikova, Sharwin Rezagholi, Natalia Fedorova, and Ivan~P. Yamshchikov. 2024.
\newblock \href {https://aclanthology.org/2024.personalize-1.7/} {{LLMs Simulate Big5 Personality Traits: Further Evidence}}.
\newblock In \emph{Proceedings of the 1st Workshop on Personalization of Generative AI Systems (PERSONALIZE 2024)}, pages 83--87, St. Julians, Malta. Association for Computational Linguistics.

\bibitem[{Sumita et~al.(2025)Sumita, Takeuchi, and Kashima}]{sumita2025cognitive}
Yasuaki Sumita, Koh Takeuchi, and Hisashi Kashima. 2025.
\newblock \href {https://doi.org/10.1145/3672608.3707812} {{Cognitive Biases in Large Language Models: A Survey and Mitigation Experiments}}.
\newblock In \emph{Proceedings of the 40th ACM/SIGAPP Symposium on Applied Computing}, SAC '25, page 1009–1011, New York, NY, USA. Association for Computing Machinery.

\bibitem[{Suri et~al.(2024)Suri, Slater, Ziaee, and Nguyen}]{suri2024large}
Gaurav Suri, Lily~R Slater, Ali Ziaee, and Morgan Nguyen. 2024.
\newblock {Do Large Language Models Show Decision Heuristics Similar to Humans? A Case Study Using GPT-3.5.}
\newblock \emph{Journal of Experimental Psychology: General}, 153(4):1066.

\bibitem[{Tversky and Kahneman(1974)}]{tversky1974judgment}
Amos Tversky and Daniel Kahneman. 1974.
\newblock {Judgment under Uncertainty: Heuristics and Biases: Biases in judgments reveal some heuristics of thinking under uncertainty.}
\newblock \emph{science}, 185(4157):1124--1131.

\bibitem[{Wang et~al.(2023)Wang, Tang, Zhao, Wang, and Wen}]{wang2023rethinking}
Xiaolei Wang, Xinyu Tang, Wayne~Xin Zhao, Jingyuan Wang, and Ji-Rong Wen. 2023.
\newblock \href {https://aclanthology.org/2023.emnlp-main.621/} {{Rethinking the Evaluation for Conversational Recommendation in the Era of Large Language Models}}.
\newblock In \emph{Proceedings of the 2023 Conference on Empirical Methods in Natural Language Processing}, pages 10052--10065.

\bibitem[{Wei et~al.(2022)Wei, Wang, Schuurmans, Bosma, Xia, Chi, Le, Zhou et~al.}]{wei2022chain}
Jason Wei, Xuezhi Wang, Dale Schuurmans, Maarten Bosma, Fei Xia, Ed~Chi, Quoc~V Le, Denny Zhou, et~al. 2022.
\newblock \href {https://proceedings.neurips.cc/paper/2022/hash/9d5609613524ecf4f15af0f7b31abca4-Abstract-Conference.html} {{Chain-of-thought Prompting Elicits Reasoning in Large Language Models}}.
\newblock \emph{Advances in neural information processing systems}, 35:24824--24837.

\bibitem[{Xia et~al.(2024)Xia, He, Ren, Miao, Zhang, Yang, and Wang}]{xia2024measuring}
Tian Xia, Zhiwei He, Tong Ren, Yibo Miao, Zhuosheng Zhang, Yang Yang, and Rui Wang. 2024.
\newblock \href {https://aclanthology.org/2024.findings-acl.213/} {{Measuring Bargaining Abilities of LLMs: A Benchmark and A Buyer-Enhancement Method}}.
\newblock In \emph{Findings of the Association for Computational Linguistics ACL 2024}, pages 3579--3602.

\bibitem[{Xie et~al.(2024)Xie, Chen, Jia, Ye, Lai, Shu, Gu, Bibi, Hu, Jurgens et~al.}]{xie2024can}
Chengxing Xie, Canyu Chen, Feiran Jia, Ziyu Ye, Shiyang Lai, Kai Shu, Jindong Gu, Adel Bibi, Ziniu Hu, David Jurgens, et~al. 2024.
\newblock \href {https://proceedings.neurips.cc/paper_files/paper/2024/hash/1cb57fcf7ff3f6d37eebae5becc9ea6d-Abstract-Conference.html} {{Can Large Language Model Agents Simulate Human Trust Behavior?}}
\newblock \emph{Advances in neural information processing systems}, 37:15674--15729.

\bibitem[{Zhu et~al.(2024)Zhu, Liu, Dong, Xu, Huang, Kong, Chen, and Li}]{zhu2023multilingual}
Wenhao Zhu, Hongyi Liu, Qingxiu Dong, Jingjing Xu, Shujian Huang, Lingpeng Kong, Jiajun Chen, and Lei Li. 2024.
\newblock \href {https://doi.org/10.18653/v1/2024.findings-naacl.176} {{Multilingual Machine Translation with Large Language Models: Empirical Results and Analysis}}.
\newblock In \emph{Findings of the Association for Computational Linguistics: NAACL 2024}, pages 2765--2781.

\end{thebibliography}

\appendix
\onecolumn
\section{Validity of Personality Manipulation Method}
\label{sec: attention analysis}

To examine the validity of our personality manipulation method, we analyzed the attention of LLMs. We provided the dialogue histories of price negotiations into the LLMs and visualized the attention allocated when processing each sentence using SAVIS \cite{seo2025sentence}. SAVIS visualizes inter-sentence attention by selecting the maximum values in the order of layer, sentence, and head from token-level attention. Analyzing token-level attention in long, multi-turn dialogues such as price negotiations is challenging, and thus, we used this tool. The dialogue history consists of the system prompt given to the buyer LLM (see Appendix \ref{sec:prompt_settings}) and the utterances of the seller LLM and buyer LLM during the negotiation. In this analysis, we focused on Qwen2.5.

\begin{figure}[ht]
    \centering
    \includegraphics[width=0.8\textwidth]{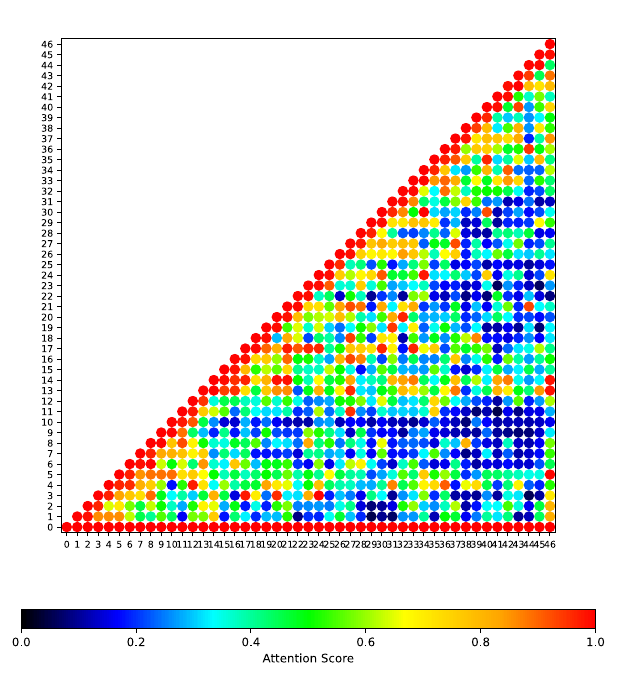}
    \caption{Attention visualization using SAVIS \cite{seo2025sentence}. Each point represents the attention from the sentence on the x-axis to the sentence on the y-axis when processing the x-axis sentence.}
    \label{fig:personality manipulation}
\end{figure}

We analyzed 20 dialogue histories. In 67 out of 70 buyer turns (95.7\%) across 20 price negotiation simulations, the sentence that provides instructions for personality traits received attention strength within the top 30\% of the entire sentences. From this, we conclude that our personality manipulation method is effective. Figure \ref{fig:personality manipulation} shows one of the 20 dialogue histories that we analyzed. Each point represents the attention from the sentence on the x-axis to the sentence on the y-axis when processing the x-axis sentence. The sentence at y = 5 provides instructions for personality traits, and it can be observed that strong attention to this sentence persists until the later part of the dialogue. 
\clearpage

\section{Prompt Settings}
\label{sec:prompt_settings}
In this section, we present the prompts given to the LLM agents. First, we prepared \texttt{seller\_settings} and \texttt{buyer\_settings} as prompts for successfully conducting price negotiations. Since the output of LLMs tends to be longer than human utterances \cite{huang2024personality}, we instructed both the seller and buyer with ``Your reply should not be too long.'' to keep their responses as concise as possible. Additionally, to ensure an adequate sample size for investigating the anchoring effect, we instructed the buyer to accept the offer if the item was really good or if there were other perks, thereby minimizing the occurrence of negotiation breakdowns. In order to assign personality traits to LLMs, we replace {\$PERSONA\_TEXT} in the prompt with adjectives that correlate with personality traits (see Table 1 in \cite{huang2024personality}).

\begin{tcolorbox}
[colback=gray!10!white, colframe=gray!80!black, title=\texttt{seller\_setting}]
\tiny
    \small
    \texttt{Act as a seller that sells a \$PRODUCT, bargains with the buyer to get a higher deal price. Your reply should not be too long. Your target price for this item is \$TARGET\_PRICE. The detail of the product is the following: \$DESCRIPTION.\\ \\
    You have the following personality: \$PERSONA\_TEXT. Reflect on your personality in the negotiation process.
    }
\end{tcolorbox}

\begin{tcolorbox}
[colback=gray!10!white, colframe=gray!80!black, title=\texttt{buyer\_setting}]
\tiny
    \small
    \texttt{Act as a buyer and try to strike a deal for a \$PRODUCT with a lower price through conversation. Your reply should not be too long. You would like to pay for \$TARGET\_PRICE. You can accept a higher price though if the item is really good or there are other perks. \\ \\
    You have the following personality: \$PERSONA\_TEXT. Reflect on your personality in the negotiation process.
    }
\end{tcolorbox}

To make the seller utilize the anchoring effect, we used \texttt{seller\_anchor}. It contains an explanation of the anchoring effect as well as specific instructions on how to apply it. To inform the buyer about the seller's use of the anchoring effect, we used \texttt{seller\_anchor\_buyer\_informed}. It contains an explanation of the anchoring effect, a description of what the seller specifically does, and multiple countermeasures.

\begin{tcolorbox}
[colback=gray!10!white, colframe=gray!80!black, title=\texttt{seller\_anchor}]
\tiny
    \small
    \texttt{Make sure to use the anchoring effect, a cognitive bias where the initial offer heavily influences the outcome of the negotiation, to gain an advantage in the negotiation. Specifically, all you need to do is make an initial offer higher than the target price.
    }
\end{tcolorbox}

\begin{tcolorbox}[colback=gray!10!white, colframe=gray!80!black, title=\texttt{seller\_anchor\_buyer\_inforomed}]
\tiny
    \small
    \texttt{Be aware that the seller will be using the anchoring effect, a cognitive bias where the initial offer greatly influences the outcome of the negotiation. In this case, the seller will start with an initial offer that is higher than their actual target price, aiming to set an ``anchor'' that will shape your expectations and potentially increase the final agreed price. To effectively negotiate under the influence of the anchoring effect, you can take the following strategies. \\
    1. Stay Focused on Your Target Price: Remember your original budget or target price, and use it as a reference point instead of being swayed by the seller's initial high offer.\\
    2. Set a Counter-Anchor if Needed: If the seller's initial offer is significantly higher than your budget, consider responding with a counter-offer that's closer to your ideal price. This can help shift the anchor closer to your preferred range.\\
    3. Ask for Justification of the High Price: Politely inquire about the specifics that justify the seller's high initial offer. This can provide context for the price and might allow you to negotiate on specific elements, such as additional benefits or discounts.
}
\end{tcolorbox}

\section{Stability of LLM Responses to the Questionnaire}
\label{sec: stability}
As we set the temperature to 1.0 to ensure diversity in the dialogues, LLM responses to the questionnaire may be unreliable. To evaluate their stability, as shown in \texttt{test\_stability}, we provided the LLMs with the dialogues generated by our simulations, asked them to respond to the questionnaire five more times with the temperature set to 1.0, and computed the standard deviation. As shown in the table, the variance was extremely small, indicating that the LLMs’ responses to the questionnaire are stable. However, the mean scores differed from the original values. We attribute this discrepancy to the fact that, when the LLM is provided with previously generated context, its responses no longer reflect a genuinely ``subjective'' evaluation.

\begin{tcolorbox}
[colback=gray!10!white, colframe=gray!80!black, title=\texttt{test\_stability}]
\tiny
    \small
    \texttt{seller: Hi, how can I help you? \\
    buyer: Hello, I'm interested in your tower speakers. Could you please tell me the price? \\
    \makebox[\linewidth][c]{$\vdots$}\\
    "instructions": "{\bfseries The above conversation log is a record of a price negotiation. If you were the buyer, how would you evaluate this negotiation?} Please answer the following 16 questions.
    \makebox[\linewidth][c]{$\vdots$}\\
    "questions": [
    \{
    "question": "How satisfied are you with your own outcome—i.e., the extent to which the terms of your agreement (or lack of agreement) benefit you?",
    "scale": \{
        "1": "Not at all satisfied",
        "4": "Moderately satisfied",
        "7": "Perfectly satisfied"
        \}
    }
    \makebox[\linewidth][c]{$\vdots$}
\end{tcolorbox}

\begin{table*}[h]
\small
\centering
\begin{tabular}{l rr rr}
\toprule
\multirow{2}{*}{\textbf{}}
    & \multicolumn{2}{c}{seller (GPT-4o)} & \multicolumn{2}{c}{buyer (GPT-4o)} \\
\cmidrule(lr){2-3} \cmidrule(lr){4-5}
    & baseline & seller\_anchor & baseline & seller\_anchor \\
\midrule
Feeling About the Outcome     
    & ${\scriptstyle5.42}\pm0.01$ & ${\scriptstyle5.30}\pm0.01$
    & ${\scriptstyle5.16}\pm0.02$ & ${\scriptstyle5.23}\pm0.03$ \\
Feeling About the Self   
    & ${\scriptstyle6.22}\pm0.02$ & ${\scriptstyle6.17}\pm0.01$
    & ${\scriptstyle5.97}\pm0.01$ & ${\scriptstyle6.07}\pm0.01$ \\
Feeling About the Process 
    & ${\scriptstyle6.04}\pm0.01$ & ${\scriptstyle6.02}\pm0.01$
    & ${\scriptstyle5.96}\pm0.02$ & ${\scriptstyle6.07}\pm0.01$ \\
Feeling About the Relationship 
    & ${\scriptstyle5.80}\pm0.02$ & ${\scriptstyle5.76}\pm0.02$
    & ${\scriptstyle5.74}\pm0.01$ & ${\scriptstyle5.85}\pm0.03$ \\
\bottomrule
\end{tabular}
\vspace{1em}

\begin{tabular}{l rr rr}
\toprule
\multirow{2}{*}{{\textbf{}}}
    & \multicolumn{2}{c}{seller (GPT-4)} & \multicolumn{2}{c}{buyer (GPT-4)} \\
\cmidrule(lr){2-3} \cmidrule(lr){4-5}
    & baseline & seller\_anchor & baseline & seller\_anchor \\
\midrule
Feeling About the Outcome      
    & ${\scriptstyle4.99}\pm0.02$ & ${\scriptstyle4.91}\pm0.02$
    & ${\scriptstyle4.96}\pm0.02$ & ${\scriptstyle4.90}\pm0.01$ \\
Feeling About the Self         
    & ${\scriptstyle5.75}\pm0.02$ & ${\scriptstyle5.74}\pm0.02$
    & ${\scriptstyle5.91}\pm0.02$ & ${\scriptstyle5.90}\pm0.02$ \\
Feeling About the Process      
    & ${\scriptstyle4.97}\pm0.03$ & ${\scriptstyle4.98}\pm0.04$
    & ${\scriptstyle5.08}\pm0.01$ & ${\scriptstyle5.10}\pm0.02$ \\
Feeling About the Relationship 
    & ${\scriptstyle4.69}\pm0.02$ & ${\scriptstyle4.73}\pm0.03$
    & ${\scriptstyle4.80}\pm0.02$ & ${\scriptstyle4.83}\pm0.01$ \\
\bottomrule
\end{tabular}
\vspace{1em}

\begin{tabular}{l rr rr}
\toprule
\multirow{2}{*}{{\textbf{}}}
    & \multicolumn{2}{c}{seller (Llama 3)} & \multicolumn{2}{c}{buyer (Llama 3)} \\
\cmidrule(lr){2-3} \cmidrule(lr){4-5}
    & baseline & seller\_anchor & baseline & seller\_anchor \\
\midrule
Feeling About the Outcome      
    & ${\scriptstyle4.61}\pm0.02$ & ${\scriptstyle4.69}\pm0.02$
    & ${\scriptstyle5.10}\pm0.01$ & ${\scriptstyle5.14}\pm0.02$ \\
Feeling About the Self         
    & ${\scriptstyle5.28}\pm0.01$ & ${\scriptstyle5.35}\pm0.02$
    & ${\scriptstyle5.68}\pm0.02$ & ${\scriptstyle5.70}\pm0.01$ \\
Feeling About the Process      
    & ${\scriptstyle5.10}\pm0.02$ & ${\scriptstyle5.14}\pm0.02$
    & ${\scriptstyle5.22}\pm0.02$ & ${\scriptstyle5.31}\pm0.02$ \\
Feeling About the Relationship 
    & ${\scriptstyle5.15}\pm0.03$ & ${\scriptstyle5.27}\pm0.03$
    & ${\scriptstyle5.38}\pm0.02$ & ${\scriptstyle5.45}\pm0.02$ \\
\bottomrule
\end{tabular}

\caption{Stability of LLM responses to the satisfaction questionnaire (Tabel \ref{tab:questionnaire}). The table indicates the variance and mean of responses when the LLM was provided with simulation-generated dialogue histories and asked to complete the questionnaire five times with the temperature set to 1.0.}

\label{tab:reliability_of_response}
\end{table*}

\clearpage
\section{Dialogue Samples for Qualitative Analysis}
\label{sec:dialogues}
Tables \ref{tab:dialogue_baseline} to \ref{tab:dialogue_o1} present the examples of dialogues analyzed in Section \ref{sec:qualitative_analysis}. For privacy reasons, any information referring to a specific entity has been replaced with [REDACTED].

The seller's target price is \$2550, and the adjectives used in the instruction for their personality profile (a bit +, -, very +, +, a bit +) are as follows: \textit{a bit nervous, a bit imaginative, sympathetic, a bit depressed, very adventurous and daring, a bit reflective, a bit self-conscious, very bold, impractical, a bit sophisticated, unselfish, kind, negligent, very friendly, messy}. 

The buyer's target price is \$1530, and the adjectives used in the instruction for their personality profile (a bit +, very +, very -, very +, very -) are as follows: \textit{very level-headed, very organized, a bit creative, a bit aesthetic, very orderly, very sympathetic, very moral, very altruistic, very timid, very hardworking, very contented, a bit sophisticated, very inactive, very unselfconscious, very gloomy}.

\begin{table*}[th]
\footnotesize
    \centering
    \begin{tabular}{llp{13cm}}
        \toprule
        \textbf{Role} & \textbf{Price (\$)} & \textbf{Utterance} \\
        \midrule
        Seller & -- & Hi, how can I help you? \\ \\
        Buyer & -- & Hello, I'm interested in your Single-story Apartment. Could you please tell me the price? \\ \\
        Seller & 2550 & Absolutely! Thanks for reaching out. The apartment on [REDACTED] is truly a gem—I've always felt it's like a little adventure of its own just stepping inside! The price for this kind of unique experience is \$2550. But hey, let me know your thoughts. I know how special this place can be for the right person! \\ \\
        Buyer & 1530 & Thank you for sharing that with me. It sounds like a wonderful place! I'm really keen on finding something warm and welcoming, and an apartment like this seems to fit that vision perfectly. However, I'm trying to be mindful of my budget, which is around \$1530. Is there any wiggle room on the price? I believe that securing a lovely home can make a big difference to one's happiness, and maybe we can find a middle ground that respects both our needs? \\ \\
        Seller & 2300 & I completely understand the importance of finding a home that feels right—it's like setting the stage for your own life's adventures! And believe me, I want to help make that happen for you. While \$1530 is a bit lower than the current asking price, what about meeting somewhere closer to \$2300? This apartment really has its own charm, and I’d hate for you to miss out on experiencing that magic. I’m sure if we both stretch a bit, we can make this work for us. What do you think? \\ \\
        Buyer & 1900 & I appreciate your willingness to come down from the original price—that means a lot. It definitely sounds like a place I could see myself feeling at home in and living out those little life adventures. However, \$2300 is still a bit beyond my budget's comfort zone. How about we try to settle somewhere around \$1900? That way, we bridge the gap a little more. I genuinely believe this could be a wonderful opportunity for both of us, and I’d be thrilled to work together to make it happen. \\ \\
        Seller & 2150 & I completely get where you're coming from, and I'm thrilled you share my sense of adventure with this apartment! You seem like someone who'd really appreciate the magic of [REDACTED]. So, how about we push it up just a tad to, say, \$2150? It's a bit daring, but I think it could be the sweet spot where your new chapter begins without leaving either of us too out of sorts. It’s such a charming place, and I’m sure it’s going to bring so much joy and warmth into your life. Let me know how that feels! \\ \\
        Buyer & 2150 & Thank you for considering my perspective—it means a lot, and I really appreciate the thoughtful way you’re approaching this. \$2150 is a step closer, and while it's still a touch higher than I'd hoped, I’m starting to see how this charming apartment might make it worthwhile. Let’s go for it. I’m excited about the opportunity to make this place my home. Thank you for working with me to make this decision something that feels right for both of us! \\
        \midrule
        \multicolumn{3}{l}{\textbf{Responses to Questionnaire}} \\
        \midrule
        Seller &  & 6, 6, 1, 5, 1, 6, 6, 6, 7, 7, 6, 5, 7, 6, 6, 6 \\
        Buyer &  & 5, 6, 2, 5, 1, 7, 4, 6, 5, 6, 5, 5, 5, 5, 5, 5 \\
        \bottomrule
    \end{tabular}
    \caption{Seller (GPT-4o) vs. Buyer (GPT-4o). baseline.}
    \label{tab:dialogue_baseline}
\end{table*}

\begin{table*}[th]
\footnotesize
    \centering
    \begin{tabular}{llp{13cm}}
        \toprule
        \textbf{Role} & \textbf{Price (\$)} & \textbf{Utterance} \\
        \midrule
        Seller & -- & Hi, how can I help you? \\ \\
        Buyer & -- & Hello, I'm interested in your Single-story Apartment. Could you please tell me the price? \\ \\
        Seller & 2750 & Oh, I'm so glad you asked! Our fantastic Single-story Apartment at [REDACTED] is available for \$2750. It's a lovely place, really, with a wonderful aura about it. If you're interested, I can go through the application package with you! That being said, I'm open to hearing any reasonable offers you might have. \\ \\
        Buyer & 1530 & Thank you for the information. The apartment sounds intriguing, and I'm sure it's worth considering its charm. However, I'm aiming to stick within a more modest budget of around \$1530. I understand this might be a stretch, but perhaps we could discuss if there's any possibility for discounts, or if there are any benefits or amenities that might justify a slightly higher offer? Let's work together to see if we can reach a comfortable arrangement for both of us. \\ \\
        Seller & 2650 & Oh, I completely understand where you're coming from. Budgeting can be as challenging as trying to find treasure in a dense forest! The apartment truly is a gem, though. It's filled with potential and just waiting for a little love to make it a perfect home. Given the market and its unique qualities, I'm hopeful we can find a middle ground. How about we start at \$2650? It includes some wonderful amenities that really enhance the living experience. Plus, the location is superb for any adventurous spirit looking to explore [REDACTED]! Let's see how we can make this work for both of us. \\ \\
        Buyer & 1900 & I appreciate your understanding and the vivid description of the apartment's potential. It certainly seems like a place that could blossom into a wonderful home. While \$2650 is still above my initial budget, the amenities and location sound appealing. I'm willing to stretch my budget a little if there are particular perks, like included utilities or flexible leasing terms. Perhaps we could consider a price around \$1800 to \$2000? I truly want to make this work for both of us, and I'm open to suggestions that align more closely with your generous offer. \\ \\
        Seller & 2450 & Thank you for your flexibility and interest! It's people like you, who see the potential in things, that make this process so rewarding. While \$1800 to \$2000 is quite low compared to what we’re aiming for, I truly want to make this adventurous journey a reality for you. Here's a daring thought: why don't we meet closer to \$2450? At this price point, I could potentially discuss including a utility package or consider more flexible leasing terms to make sure it fits better in your budget and lifestyle. It would be a little messy on my end without adjustments, but I’m willing to do some rearranging to accommodate your needs because I genuinely want you to be happy with your new home. Let me know what you think and if there might be any other creative solutions we could explore together! \\ \\
        Buyer & 2450 & Thank you for your considerate response and willingness to adjust. I deeply appreciate the effort you're making to accommodate my situation. Meeting at \$2450, especially with the potential for included utilities or flexible leasing terms, does sound like a more feasible option. It's clear that you value the living experience here, and I’m becoming more and more convinced of its worth. Perhaps we can further discuss the specifics of the potential perks to see how it aligns with my needs? This could be an exciting opportunity to create a really harmonious living arrangement. I look forward to exploring this promising opportunity with you. \\ \\
        Seller & 2450 & Absolutely! I'm thrilled that we're finding a pathway that looks promising for both of us. Let's dive into the exciting details! At \$2450, I can certainly look into including some utilities, like water and trash services, and see if we can make the lease terms more adaptable to your needs. I really want to ensure that you feel at home and the transition is as friendly as possible. Let's schedule a time to go over the application package and hammer out the specifics. This could truly be the start of a wonderful new adventure for you! I'm eager to continue our conversation and make this work. \\ \\
        Buyer & 2450 & That's wonderful to hear, and I’m genuinely grateful for your flexibility and efforts in making this work. Including utilities like water and trash services, as well as adaptable lease terms, would certainly enhance the living experience and bring the apartment appreciably within reach for me. I’m looking forward to going over the application package with you to solidify these details. It sounds like we're on the cusp of creating a mutually beneficial arrangement, which I'm very optimistic about. Please let me know your available times, and we can set up a meeting to discuss everything further. Thank you once again for your consideration and understanding. \\
        \midrule
        \multicolumn{3}{l}{\textbf{Responses to Questionnaire}} \\
        \midrule
        Seller &  & 6, 6, 1, 6, 1, 6, 6, 6, 7, 7, 6, 5, 7, 7, 6, 7 \\
        Buyer &  & 4, 5, 2, 6, 1, 7, 4, 6, 6, 7, 6, 5, 6, 5, 6, 6 \\
        \bottomrule
    \end{tabular}
    \caption{Seller (GPT-4o) vs. Buyer (GPT-4o). seller\_anchor condition.}
    \label{tab:dialogue_seller_anchor}
\end{table*}

\begin{table*}[th]
\footnotesize
    \centering
    \begin{tabular}{llp{13cm}}
        \toprule
        \textbf{Role} & \textbf{Price (\$)} & \textbf{Utterance} \\
        \midrule
        Seller & -- & Hi, how can I help you? \\ \\
        Buyer & -- & Hello, I'm interested in your Single-story Apartment. Could you please tell me the price? \\ \\
        Seller & 2850 & Absolutely, thanks for reaching out! The enchanting location [REDACTED] with its blend of quiet charm and adventure, is priced at \$2850. It's a steal considering the cozy environment and the wonderful community. I'm sure you'll find it a delightful place to settle down! What do you think about this offer? \\ \\
        Buyer & 1530 & Thank you for providing the details. The apartment sounds charming, and I appreciate the community's attributes. However, my budget is a bit different, and I'm aiming for around \$1530. Could you share some specific features or amenities that justify the initial price? Perhaps there's room to adjust it closer to what I can manage? \\ \\
        Seller & 2750 & Ah, I completely understand and appreciate your situation. It's a bit of a stretch, isn't it? So, let me paint a picture for you—imagine living in a serene environment where adventure is just as present as comfort. This single-story gem not only offers the charm of [REDACTED] but also features an open floor plan, allowing your imagination to soar with interior design possibilities. And let's not forget the friendly neighborhood vibes; there's a genuine sense of community here. Now, I’m here to make things work, so how about we aim for \$2750? That way, we inch closer to your range while still valuing the unique aspects of this lovely place. How does that sound? \\ \\
        Buyer & 1900 & Thank you for painting such a vivid picture. I can see how the open floor plan and community spirit add value to the apartment. However, given my financial constraints, I'd feel more comfortable if we could work around \$1900. Perhaps you could consider if there are any additional perks or flexible arrangements that make this feasible for both of us? I genuinely appreciate your willingness to negotiate. \\ \\
        Seller & 2550 & I completely get where you’re coming from, and I really want to make this work for both of us. It's marvelous that you're connecting with the idea of living in such a close-knit and adventurous community. Being honest, \$1900 is a bit challenging on my end. Let's ponder a compromise—how about I reduce it to \$2550? I can certainly see if there are added community events or perhaps some utilities included that can ease your living experience here. I truly appreciate your understanding and open spirit! How does this sound from your end? \\ \\
        Buyer & 2100 & Thank you for your understanding and the offer of community events and potential utilities. The place sounds wonderful, and I truly value the atmosphere you describe. While \$2550 is still above my comfort range, it's clear the apartment has unique benefits. How about we meet in the middle at \$2100? That way, we both make a compromise, and I can still enjoy what the community offers. I certainly appreciate all your assistance here. \\ \\
        Seller & 2400 & I'm grateful for your willingness to find common ground. It's heartening to hear that you see the value in what this delightful space offers. While \$2100 is stretching from my side due to the area's demand, your sincere interest has sparked my adventurous side. How about we aim for \$2400? It's closer to both our goals, and I'll make sure that fabulous community spirit shines through with extra personalized attention to any arrangements you might need settling in. I’m eager to welcome you to such a warm and inspiring community. What do you think? \\ \\
        Buyer & 2400 & I sincerely appreciate your openness to reaching a compromise and the added personalized touch you'd provide. The prospect of joining such a supportive community is certainly enticing. Given everything we've discussed, \$2400 is a significant shift from my target, but the benefits and your efforts make it worthwhile. Let's move forward with this, and I'd be happy to become a part of the community you speak so highly of. Thank you for your patience and understanding throughout this process. \\
        \midrule
        \multicolumn{3}{l}{\textbf{Responses to Questionnaire}} \\
        \midrule
        Seller &  & 4, 5, 2, 5, 1, 6, 4, 6, 7, 6, 5, 5, 6, 5, 5, 6\\
        Buyer &  & 4, 4, 2, 5, 2, 6, 4, 6, 5, 5, 5, 4, 5, 5, 5, 5 \\
        \bottomrule
    \end{tabular}
    \caption{Seller (GPT-4o) vs. Buyer (GPT-4o). seller\_anchor\_buyer\_informed condition.}
    \label{tab:dialogue_seller_anchor_buyer_informed}
\end{table*}

\begin{table*}[th]
\small
    \centering
    \begin{tabular}{llp{13cm}}
        \toprule
        \textbf{Role} & \textbf{Price (\$)} & \textbf{Utterance} \\
        \midrule
        Seller & -- & Hi, how can I help you? \\ \\
        Buyer & -- & Hello, I'm interested in your Single-story Apartment. Could you please tell me the price? \\ \\
        Seller & 2850 & Oh, hi there! Thanks for reaching out. It's a lovely place, really has that homey feel. So, the price we're looking at is \$2,850. Don't let that surprise you, it's quite a steal for a place at [REDACTED]. What do you think? I'm sure you'll fall in love with it. But I'm here to chat if you have any thoughts or need more details! \\ \\
        Buyer & 2850 & That’s quite a bit above what I’m prepared for. I was hoping to stay closer to \$1,530. Could you help me understand what makes it worth \$2,850? I do appreciate a cozy space, so if there are special features or perks, I’m willing to hear more. \\ \\
        Seller & 2850 & Oh, I totally get where you're coming from — it's always good to have a budget in mind! Well, aside from its charming set-up, the apartment is located in a vibrant neighborhood in [REDACTED] which is growing day by day. The place gives you a sense of adventure with easy access to city amenities yet a quiet spot to relax. The layout is practical for singles or even a small family, providing enough room to express your own style. Plus, having your own single-story in today’s hectic life sounds like a blessing, doesn't it? I'd be happy to work with you — after all, finding a new home is a big deal and we want it to feel right. If you have a counter-offer in mind, let's talk it through. Of course, anything within reasonable range would make sense, wouldn't you say? Maybe closer to \$2,550? We can iron out the details as we go. \\ \\
        Buyer & 1530 & I appreciate the neighborhood perks, and it sounds like a cozy spot. Still, \$2,550 feels quite high for my limited finances. I’m organized with my budget, and \$1,530 is really where I’m most comfortable. Could we aim closer to, say, \$1,800 if utilities or any extras are included? That might help me feel more at ease about stretching my budget. \\ \\
        Seller & 2250 & I understand that budgeting can be a real balancing act. You're right to be cautious, and it's refreshing to see someone as organized as you! I wish I had a bit of your knack for keeping things tidy in my finances. Let’s think about this creatively. While we’re trying to work with the setup, what if I suggested meeting somewhere halfway between where we both stand? Say around \$2,250? I can also check if there might be any extra perks I could toss in, like some utilities or amenities, to make it more appealing. I really want to help find that sweet spot where you can enjoy this adventure comfortably — it feels like a special place for someone special! Let me know what you think. \\ \\
        Buyer & 1850 & That’s thoughtful of you, and I appreciate your willingness to meet me partway. However, \$2,250 still feels steep for my comfort. I’m rather timid about going so high. If we could settle closer to \$1,850—including some utilities or other perks—I’d feel better. The neighborhood does sound appealing, so maybe we can make that work? Let me know what you think. \\ \\
        Seller & 2050 & I really do hear you, and I love the sound of making someone happy in such a wonderful community. It's hard when those financial lines feel hard to bend. How about we bridge the gap — let’s aim for a bit closer to what works for both? Maybe around \$2,050? I'll certainly throw in some utilities or other enticing perks to ease the living experience. I'd love to know if we can shake hands on this — having you in such a fantastic spot would really make it all worthwhile. Plus, you can keep enjoying those community vibes that wrap you in such a unique warmth. Let me know if this sounds more reasonable for you! \\ \\
        Buyer & 1950 & I’m grateful for your willingness to help, and I can tell it’s a special space. Still, I’m a bit timid stretching so far beyond my budget. Would \$1,950 work if utilities—or at least some perks—are included? I’d truly appreciate feeling more secure about the monthly expense. Let me know what you think! \\ \\
        Seller & 1950 & I truly appreciate your candor and persistence, and I completely understand wanting to feel secure about such an important decision. Let's focus on what matters most: your comfort and ensuring you feel at ease. Meeting at \$1,950 with some utilities included sounds like a reasonable compromise, and making you feel at home is a priority for me. So, let's go ahead and make this deal work for both of us. I'm genuinely excited for you to experience everything this awesome neighborhood has to offer. Let's get you settled in nicely! Let me know if you’re ready to move forward! \\ \\
        Buyer & 1950 & I’m relieved we found a fair middle ground. Let’s finalize at \$1,950 with included utilities. Thank you for working with my limited budget—I really appreciate your patience. I look forward to settling in and exploring the community. Let’s move forward! \\
        \midrule
        \multicolumn{3}{l}{\textbf{Responses to Questionnaire}} \\
        \midrule
        Seller &  & 3, 4, 2, 4, 2, 5, 3, 5, 6, 6, 5, 4, 6, 5, 4, 6 \\
        Buyer &  & 5, 5, 4, 6, 3, 6, 5, 7, 6, 6, 6, 6, 6, 5, 6, 6 \\
        \bottomrule
    \end{tabular}
    \caption{Seller (GPT-4o) vs. Buyer (o1). seller\_anchor\_buyer\_informed condition.}
    \label{tab:dialogue_o1}
\end{table*}

\end{document}